# A self-supervised learning-based 6-DOF grasp planning method for manipulator

Gang Peng[1,2], Zhenyu Ren[1,2], Hao Wang[1,2,*], Xinde Li[3]

1. School of Artificial Intelligence and Automation, Huazhong University of Science and Technology, Wuhan 430074, China

2. Key Laboratory of Image Processing and Intelligent Control, Ministry of Education, Wuhan 430074, China

3. IEEE senior member, School of Automation, South East University, Nanjing, China

Abstract: To realize a robust robotic grasping system for unknown objects in an unstructured environment, large amounts of grasp data and 3D model data for the object are required, the sizes of which directly affect the rate of successful grasps. To reduce the time cost of data acquisition and labeling and increase the rate of successful grasps, we developed a self-supervised learning mechanism to control grasp tasks performed by manipulators. First, a manipulator automatically collects the point cloud for the objects from multiple perspectives to increase the efficiency of data acquisition. The complete point cloud for the objects is obtained by utilizing the hand-eye vision of the manipulator, and the TSDF algorithm. Then, the point cloud data for the objects is used to generate a series of six-degrees-of-freedom grasp poses, and the force-closure decision algorithm is used to add the grasp quality label to each grasp pose to realize the automatic labeling of grasp data. Finally, the point cloud in the gripper closing area corresponding to each grasp pose is obtained; it is then used to train the grasp-quality classification model for the manipulator. The results of data acquisition experiments demonstrate that the proposed method allows high-quality data to be obtained. The simulated results prove the effectiveness of the proposed grasp-data acquisition method. The results of performing actual grasping experiments demonstrate that the proposed self-supervised learning method can increase the rate of successful grasps for the manipulator.

Key Words: Manipulator, Grasping, Self-Supervised Learning, 3D Reconstruction

## 1. INTRODUCTION

In the field of robotics, research on grasping unknown objects in unstructured environments is very popular; in particular, grasp planning based on visual information is essential to the development of intelligent robots. Nevertheless, there are still some challenges that must be overcome: 1) In a scene where multiple objects are stacked, the time to determine a feasible grasping posture is undesirably lengthy; 2) The geometric shapes of possible target objects are diverse and irregular, and a large amount of grasp data is required to train the algorithm; 3) If the robustness of grasp planning algorithm is insufficient, the change of working environment will lead to a significant decline in the success rate of grasp.

Conventional algorithms for grasp models use a template matching algorithm and 3D model of the object to accurately calculate the pose of the object, and then perform the grasp. Kehoe et al. [1] first constructed a 3D model of the target object, and then used the Graspit! toolkit to build a grasp database for their model. With their method, a series of object models and corresponding feasible grasps are stored in the database. Then, the database is searched for a template that matches the point cloud for the target object; finally, the target object is grasped according to the preset grasp method. Owing to steady advancements in deep learning technology, Xiang et al. proposed the end-to-end object pose estimation network PoseCNN [2], which uses three neural network branches to realize object pose estimation; it also enhanced the robustness and increased the accuracy of object position estimation algorithms, which can be used to guide the trajectory of grasp-purposed manipulators. The two above-mentioned methods rely on 3D model data for the target object and are still limited in their robustness and ability to enable real-time adjustments; moreover, they cannot be applied to grasp unknown objects.

Model-free grasp optimization algorithms focus on using the geometric information contained in the object's 3D point cloud to perform grasp planning and do not require accurate estimation of the object's pose. Regarding the robotic grasp process, this type of method does not require an accurate 3D model of the target object, which can be used to generate high-quality grasp poses for unknown objects. An example of model-free object grasp technology is the grasp pose detection (GPD) method proposed by Gualtieri et al. [3], which generates a series of candidate grasp poses by using the 3D point cloud of the object

This work was supported by the National Natural Science Foundation of China (No. 91748106), and Hubei Province Natural Science Foundation of China (No. 2019CFB526).

* Corresponding author, Email: wa_haoo@hust.edu.cn

Gang Peng, PhD, Assoc. Prof, IEEE member, Email: penggang@hust.edu.cn; Zhenyu Ren (Co-First Author) Master, Email: renzhenyu@hust.edu.cn; Hao Wang (Corresponding Author), Master degree candidate, Email: wa_haoo@hust.edu.cn; Xinde Li, PhD, Prof, IEEE senior member, Email: xindeli@seu.edu.cn.

and geometric information on the parallel two-fingered gripper at the end of the manipulator and creates classification labels through the implementation of a force-closure analysis algorithm; the grasp pose quality is then classified by using a convolutional neural network (CNN). With the development of cloud computing and big data technology, Mahler et al. proposed the Dex-Net [4, 5] series of datasets and related algorithms to enable robust grasp planning for the manipulator. Dex-Net researchers collected 10,000 independent 3D object models and used grasp wrench space analysis to create grasp pose classification labels. They then employed cloud computing technology to train a grasp-quality CNN, which ultimately yielded a robust grasp pose classification model. CNNs are employed to classify the grasp quality in both of the above-mentioned methods to realize the grasping of unknown objects; thus, large-scale objects and grasp data are required to ensure the robustness of the trained grasp quality classification model.

From the above analysis of existing grasp planning methods for manipulators, it can be ascertained that, irrespective of whether it is a model-free algorithm, large-scale objects and grasp data are required to realize robust manipulator grasping performance. To overcome the challenges related to obtaining the required data in the grasp scene, improve autonomous learning ability, and increase the rate of successful grasps for the manipulator, we developed a six-degrees-of-freedom (6-DOF) grasp planning method with self-supervised learning. Our contributions to the field can be summarized as follows:

1) The self-supervised learning mechanism allows the point cloud data for objects to be obtained, and it automatically labels the grasp data without requiring the manipulator to execute the actual grasp task; this reduces the time cost of data acquisition and labeling, and increases the level of automation in the grasping task;

2) The complete point cloud for target objects is obtained by using the hand-eye vision of the manipulator and implementing a truncated signed distance function (TSDF) algorithm. A quality evaluation index for the object point cloud data has been established, and an object point cloud dataset for real multi-target scenes has been created;

3) The implemented force-closure decision algorithm adds the grasp quality label to each 6-DOF grasp pose to enable automatic labeling of grasp data; this enables rapid creation of the grasp dataset.

## 2. PROBLEM STATEMENT

In our proposed method, the input of the grasp planning module is the single-view point cloud for the object; the feasible final 6-DOF grasp pose is obtained by means of grasp pose sampling and classification. If an unknown object $O$ is given, the coefficient of friction $\mu \in \mathbb{R}$ between the object and gripper, the object's geometric and mass properties $M_o$, and the object's 6-DOF pose $W_o \in \mathbb{R}^6$ are related to the grasp. Thus, $s_o = (M_o, \mu, W_o)$ can be used to represent the state of the object. The task of the manipulator grasp planning module is to find a feasible final 6-DOF grasp pose $g = (p, r) \in \mathbb{R}^6$, where $p = (x, y, z) \in \mathbb{R}^3$ and $r = (r_x, r_y, r_z) \in \mathbb{R}^3$ respectively specify the position and orientation of the grasp $g$.

It is assumed that the complete point cloud $PC \in \mathbb{R}^{3*N}$ containing $N$ points for the object can be collected by using a 3D scanner. Furthermore, to evaluate the quality of the final 6-DOF grasp pose $g$, the metric $Q(s, g, \mu) \in \mathbb{R}$ is used to represent the grasp quality. Based on the above-described assumptions and definitions, the relationship between the grasp pose and complete point cloud for the object can be described as is shown in Figure 1.

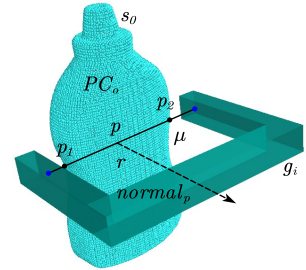

Figure 1. Relationship between the grasp pose and complete point cloud for the object.

The state information $s_o$ of the object $O$ is hidden in the complete point cloud $PC_o$ for the object $O$. After sampling, the grasp pose set $G=\{g_1, g_2 \cdots g_i\} | i \in \mathbb{N}$ can be obtained; then, the quality metric $Q_i$ can be calculated by using the coefficient of friction $\mu$ between the object and grasp pose $g_i$. According to $Q_i$, the elements in the grasp pose set $G$ can be sorted and subsequently used to guide the manipulator to execute grasp actions.

Depth cameras, which can directly obtain the 2.5D single-view point cloud data for an object, are typically employed as vision sensors in actual manipulator grasp tasks. Because it is difficult to obtain the complete state information $s_o$ for the target object, it is impossible to calculate the grasp quality metric $Q(s, g, \mu)$. To solve this problem, it is necessary to learn a new quality metric $Q_\gamma(P, g) \in \{c_0, c_1, \cdots\}$ through the process of training the model. This quality metric can only be calculated by using the single-view point cloud $P$ for the object, and the corresponding grasp pose $g$; $\gamma$ is a learning-based classification model of grasp quality, and $c_0, c_1, \cdots$ is a series of labels that characterize the quality of grasp pose $g$.

However, model training requires a large amount of 3D point cloud data for the object, as well as the corresponding 6-DOF grasp data and classification labels. Thus, to grasp unknown objects in an unstructured environment, there are three key obstacles to overcome: 1) How to obtain a large amount of point cloud data for objects in a real grasp scene; 2) How to generate classification labels for each grasp pose; and 3) How to design a grasp quality classification model.

To overcome the above-described obstacles, we constructed a self-supervised learning-based 6-DOF grasp planning algorithm framework for manipulators; it consists of two sub-modules, i.e., data acquisition and grasp planning sub-modules, as shown in Figure 2. The grasp planning sub-module is responsible for generating a series of candidate grasp poses by using the single-view point cloud, classifying and scoring the grasp poses by using a grasp quality classification neural network model, and providing optimal grasp recommendations for the manipulator based on the grasp pose scores. The data acquisition sub-module obtains the complete point cloud for the target object by performing desktop-level 3D reconstruction; it then implements a force-closure decision algorithm to analyze the quality of the grasp pose to realize the automated acquisition and labeling of multi-target grasp data in a real grasp scene. This sub-module also provides the training data that allows the grasp planning sub-module to learn a new quality metric $Q_\gamma(P, g)$.

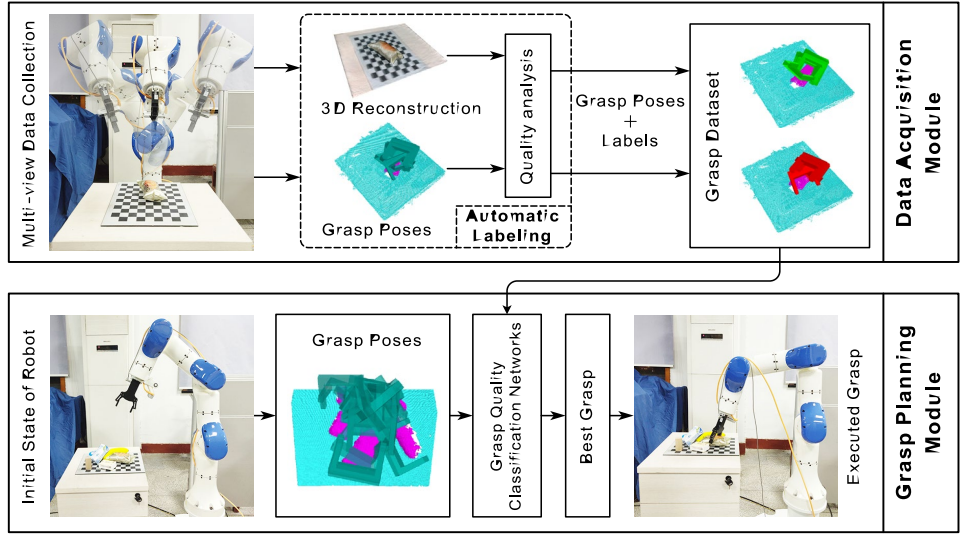

Figure 2. Framework of proposed self-supervised learning-based 6-DOF grasp planning algorithm for manipulators.

## 3. GRASP DATA ACQUISITION

### 3.1 Desktop-Level 3D Reconstruction

In the proposed model, 3D reconstruction is an important part of the self-supervised learning-based data acquisition sub-module. Unlike large-scale 3D reconstruction algorithms such as BundleFusion[6], the 3D reconstruction algorithm proposed in this paper is purposed for desktop-level manipulator grasp scenes; additionally, the scene area to be reconstructed should be small to realize highly accurate reconstruction.

In this model, the base coordinate system of the manipulator is the world coordinate system $O_{world}$; thus, the relationship between the end coordinate system $O_{end}$ and base coordinate system $O_{base}$ is ${}^b_e T = {}^w_e T$, which can be calculated by solving the forward kinematics equation for the manipulator. Then, the transformation matrix ${}^c_w T$ necessary to transform the world coordinate system to the camera coordinate system can be obtained as follows:

$$ {}^w_c T = {}^w_e T \, {}^e_c T \qquad (1) $$

In the above formula, ${}^e_c T$ represents the transformation between the camera coordinate system $O_{camera}$ and end coordinate system $O_{end}$, i.e., the hand-eye transformation matrix, which can be obtained by applying hand-eye calibration to the manipulator. Using a camera pinhole imaging model, the relationship between a point $p(u,v)$ in the depth image coordinate system and the corresponding point $P(X_w, Y_w, Z_w)$ in the world coordinate system can be described as

$$ Z_c \begin{bmatrix} u \\ v \\ 1 \end{bmatrix} = \begin{bmatrix} f_x & 0 & u_0 & 0 \\ 0 & f_y & v_0 & 0 \\ 0 & 0 & 1 & 0 \end{bmatrix} ({}^w_e T \, {}^e_c T)^{-1} \begin{bmatrix} X_w \\ Y_w \\ Z_w \\ 1 \end{bmatrix} \qquad (2) $$

In the above formula, $f_x$, $f_y$, $u_0$, and $v_0$ are the inherent properties of the camera, which can be obtained by calibrating the camera's internal parameters; $Z_c$ is the depth value at point $p$.

Our method entails first collecting the depth image set $\{img_1, img_2, \cdots img_N\}$ for the scene from N perspectives, and then calculating the camera pose

$pose = {}^{c}_{w}T$ for each image frame to obtain the camera pose set $\{pose_1, pose_2, \cdots pose_N\}$. Then, Equation (2) is applied to convert the depth image data for each frame into a 3D point cloud $PC_i$; simultaneously, an improved TSDF algorithm [7] is used for point cloud data fusion to obtain the complete point cloud $PC$.

As a result of installing the camera at the end of the manipulator, and applying the forward kinematics and hand-eye transformation matrix for the manipulator to calculate the real-time pose of the camera, the time-expensive point cloud matching process that is necessary in conventional large-scale 3D reconstruction algorithms is no longer necessary; moreover, the camera pose estimation accuracy can be increased, thus increasing the dimensional accuracy of the reconstructed point cloud to allow the reconstructed complete point cloud to more accurately express the geometric information that describes objects in the scene.

### 3.2 Automated Grasp Pose Classification

In the process of grasp data acquisition, $Q_{fc} = Q(s, g, \mu)$ is applied as the grasp quality metric to realize automated grasp pose classification. This quality metric is calculated based on the mechanical relationship between the target object and gripper. The calculation of this mechanical relationship is called force-closure analysis. "Force closure" implies that the gripper, by means of making contact with the target object, can apply a force to the object to counteract other forces acting on the object, thereby ensuring that there will be no slippage at the contact point when the object is grasped.

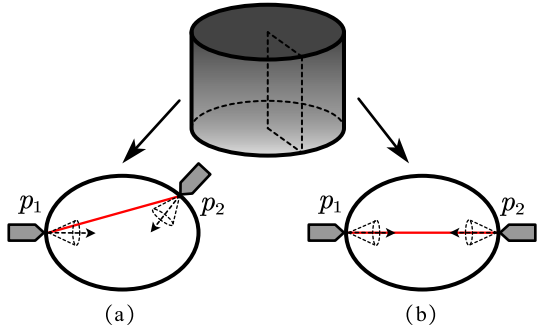

Figure 3. Schematic illustration of grasp force closure.

Because object grasping with parallel two-finger grippers involves only two contact points, according to the Nguyen theorem proposed in [8], the necessary and sufficient condition for force closure is that the line connecting the contact points should be in the friction cone at the two contact points at the same time. In Figure 3(a), the line connecting the two contact points is inside the friction cone at point $p_1$ but not inside the friction cone at point $p_2$; thus, the grasping action does not satisfy the condition for force closure. In Figure 3(b), the line connecting the two contact points passes through the centers of both friction cones; thus, this grasping action satisfies the condition for force closure.

Given two contact points $p_1$ and $p_2$ in a 3D space, the surface normal vectors $n_1$ and $n_2$ at the two contact points and the static friction coefficient $\mu$ can be obtained. Then, the unit vector from point $p_1$ to point $p_2$ can be expressed as

$$\hat{v} = \frac{p_2 - p_1}{\|p_2 - p_1\|_2} \mid p_1, p_2 \in \mathbb{R}^3 \quad (3)$$

The angle between $n_1$ and $\hat{v}$ is $\alpha_1 = cos^{-1}(n_1 * (-\hat{v}))$, and the angle between $n_2$ and $\hat{v}$ is $\alpha_2 = cos^{-1}(n_2 * (\hat{v}))$. The half-vertex angle of the friction cone at the contact point is $\beta = tan^{-1}(\mu)$; according to the Coulomb friction model, when $\alpha_1 < \beta$ and $\alpha_2 < \beta$, the force-closure condition is satisfied, and $Q_{fc} = 1$; otherwise, $Q_{fc} = 0$.

We applied the scoring mechanism proposed in [9] to improve the classification scheme for the grasp pose. For a certain grasp pose $g$, and beginning at a value of 3.0, the coefficient of friction $\mu$ is gradually reduced until $g$ does not satisfy the force-closure condition; the smallest coefficient of friction $\mu$ that satisfies the force-closure condition is recorded as the score for $g$. Figure 4(a) shows the grasp poses that satisfy the force-closure condition when $\mu = 0.4$, and Figure 4(b) shows the grasp poses that satisfy the force-closure condition when $\mu = 2.0$. It can be seen that a smaller coefficient of friction corresponds to a higher quality of grasp poses that satisfy the force-closure condition.

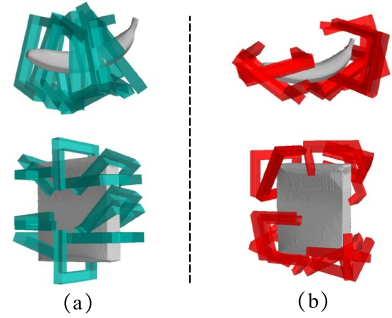

Figure 4. Illustration of grasp pose classification.

### 3.3 Self-Supervised Learning Mechanism

The grasp dataset should not only contain object point cloud data but also a series of grasp poses and corresponding labels. However, it is inefficient to employ the manipulator in actual grasping tasks for the purpose of adding classification labels for grasp poses. Thus, to realize automated annotation of manipulator grasp poses, it is best to apply the force-closure condition to analyze the reliability of the grasp pose; however, the force-closure condition can only be established when the complete geometry of the object and the positions of the contact points are known; this means that force-closure analysis for the grasp pose can only be performed when the point cloud for the object is complete.

At present, the majority of the most frequently employed 6-DOF motion planning methods for manipulator grasping require object datasets such as BigBIRD [10] and YCB [11], which include a series of single-view and complete point clouds for objects; additionally, 3D scanners and professional point cloud acquisition systems are employed in the production process. Nevertheless, these object datasets have the following problems: 1) The amount of data is limited; 2) The types of objects are limited; 3) The objects are not in a real grasp scene; and 4) Only the point cloud data for individual objects are included, even though there are typically multiple objects in the grasp scene.

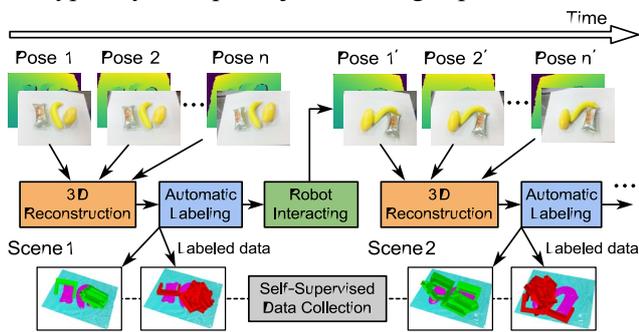

Figure 5. Overview of the proposed self-supervised data acquisition sub-module.

To overcome the challenges related to collecting grasp data in real scenes, and inspired by the work of [12, 13] and others, we designed a self-supervised learning-based data acquisition sub-module; it is shown in Figure 5. First, the initial state of the objects on the desktop is manually set; then, the manipulator automatically collects image data, performs desktop-level 3D reconstruction, and autonomously pushes the objects on the desktop according to the reconstructed scene information; data collection and 3D reconstruction of the next scene is continued until a preset amount of grasp data is collected.

Our proposed self-supervised learning-based data acquisition method entails implementation of a desktop-level 3D reconstruction algorithm to obtain the complete point cloud for objects, followed by the realization of the automatic annotation of grasp poses through the implementation of a force-closure analysis algorithm. This method has the following advantages: 1) The 6-DOF grasp planning method for the manipulator is not dependent on existing object datasets; 2) Grasp data can be collected from multi-object and stacked object scenes; 3) Data can be automatically collected and labeled in a real capture scene, thereby allowing the manipulator to learn autonomously.

## 4. DEEP LEARNING-BASED GRASP QUALITY CLASSIFICATION

The quality of the grasp pose is determined by the mechanical relationship between it and the target object. We regard the point cloud information in the closing area of the gripper as a representation of the mechanical relationship between the grasp pose and the object; it is employed as the input of the grasp quality model $\gamma$. Following input, the classification problem of grasp pose quality is transformed into a classification problem of the point cloud for the closed area of the gripper. The method of obtaining the point cloud for the closing area of the gripper is illustrated in Figure 6, where green indicates a feasible grasp, red indicates an infeasible grasp, light yellow indicates the closing area, and purple indicates the point cloud for the closing area.

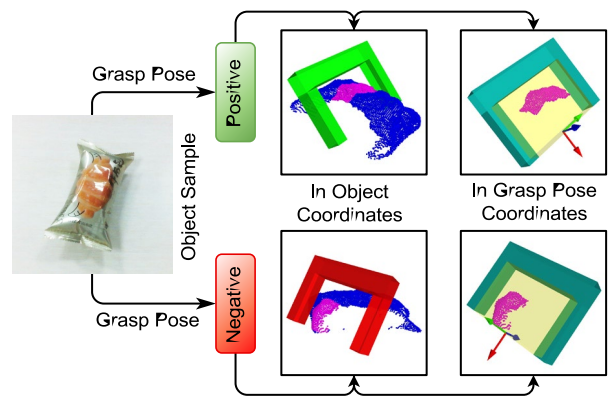

Figure 6. Method for obtaining the point cloud for the gripper closing area.

Several deep learning models for point cloud analysis that are based on PointNet [14] take the original point cloud as the input; most of them have excellent feature-extraction capability and can learn the point cloud feature information through training. This typically affords superior robustness and a very high forward propagation speed; thus, this approach is advantageous for tasks that require classification of sparse point clouds or missing point clouds. The spatial domain of the point cloud in the gripper closing area is small. Thus, applying deep learning to classify the point cloud for the gripper closing area can solve problems related to force-closure analysis not being applicable for the classification of single-view point clouds; moreover, this approach can ensure the speed and accuracy of point cloud classification, thus ensuring a high speed and success rate of the manipulator grasp planning task.

The proposed method of deep learning-based point cloud analysis for grasp quality classification is shown in Figure 7. First, the point cloud for the closing area of the gripper in the grasp pose coordinate system is obtained; then, the point cloud is taken as the input for the point cloud classification networks; finally, the scores for each category within this point cloud are output. The classification score of the point cloud for the gripper closing area can be used to determine whether the grasp pose associated with this point cloud is reliable, and the grasp pose can be sorted according to this score. Finally, the grasp pose with the highest score is selected for initiation by the manipulator for the actual grasping task.

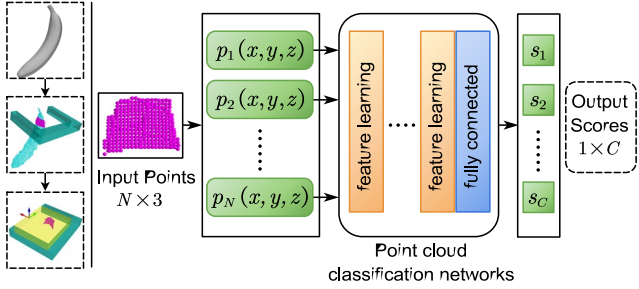

Figure 7. Grasp quality classification scheme based on deep learning-based point cloud analysis.

## 5. EXPERIMENTAL RESULTS AND ANALYSIS

To verify the effectiveness and advantages of the proposed method, a real-scene data acquisition experiment was carried out; then, a simulated grasp pose classification experiment was carried out; finally, a real-scene grasp experiment was carried out. The computer used for these experiments was configured as follows: Intel Core i5-8300H CPU, 2.3 GHz, GTX1060 GPU, and 16 GB of RAM.

We used a self-developed 7-DOF non-biased S-R-S manipulator for the real-scene data collection and grasp experiments. The end of the robot was equipped with a parallel two-finger gripper and an Intel Realsense D435i depth camera. The maximum distance between the two fingers of the gripper was approximately 7 cm.

We selected 18 common objects for the experiments; they were divided into the following three categories: fruits, boxes, and columns. The left photograph in Figure 8 shows the objects used for dataset production, and the right photograph shows the unknown objects used for grasping. Because the operational range of the gripper was limited, the sizes of the selected objects also had to be small; this condition ensured that each object had a clampable part with a width that ranged between 1 and 6 cm.

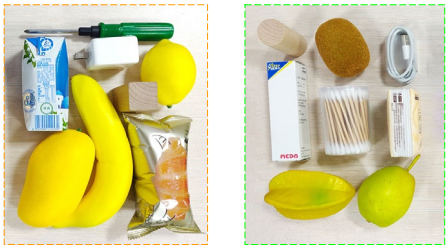

Figure 8. Objects used for dataset production and actual grasping.

### 5.1 Data Acquisition Experiment

Our grasp planning method analyzes the quality of the grasp pose by utilizing the geometric information on the object; thus, it is very important to ensure consistency between the geometric features of the single-view point cloud and the complete point cloud for the object dataset, which is guaranteed by the accuracy of the 3D reconstruction obtained via the data acquisition process.

We conducted a data acquisition experiment to 1) assess the degree of similarity between the single-view point cloud for the object collected via this method, and the complete point cloud; and 2) evaluate the quality of the obtained object data. In the experiment, a total of 450 single-view point clouds and corresponding complete point clouds were collected for the nine objects shown in the left photograph of Figure 8; various combinations of these objects were applied, and 900 single-view point clouds and corresponding complete point clouds for these combinations were collected and added to a self-supervised grasp (SSG) dataset.

To ensure the accuracy and real-time performance of the camera pose calculations, an ROS bag tool was used to save depth images and coordinate system information as the manipulator moved in real time. After the manipulator motion was completed, the bag file was parsed to obtain the coordinate system information for the camera pose calculations, and the corresponding depth images were obtained by referencing the timestamps; unmatched depth images were discarded. Finally, a series of acquired depth images and corresponding camera pose data were used to perform desktop-level 3D reconstruction. Figure 9 shows the experimental results of implementing the proposed desktop-level 3D reconstruction method.

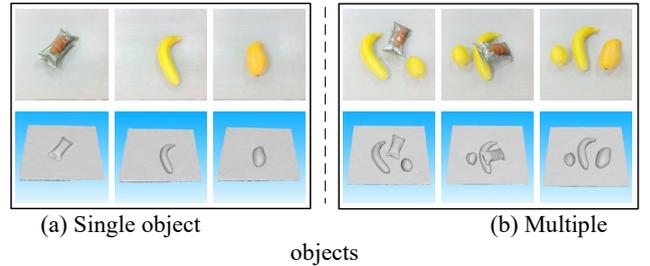

(a) Single object             (b) Multiple objects

Figure 9. Desktop-level 3D reconstruction results.

Ideally, the single-view point cloud for an object should be a subset of its complete point cloud. To quantitatively evaluate the quality of the object data in our SSG dataset, an object data precision index $acc_{Match}$ was defined to quantify the degree of similarity between the single-view point cloud and complete point cloud for the objects in the dataset.

$$acc_{Match} = \frac{point\_num_{Match}}{point\_num_{All}} \quad (4)$$

We have defined the accuracy of object data as the ratio of the number of matched points between a single-view point cloud and the corresponding complete point cloud, to the total number of points in the single-view point cloud, where the single-view point cloud and complete point cloud are represented in the same reference coordinate system. Ideally, all points in the single-view point cloud should be able to find matching points in the complete point cloud; thus, a larger value indicates higher reconstruction accuracy. We found the matching points by creating and implementing a KD-tree index for the complete point cloud, and then by traversing every point in the single-view point cloud. If at least one point in the complete point cloud could be found within a sphere with a radius $r$ at a certain point, this point was determined to be a matched point.

Regarding the YCB dataset, the complete point cloud was obtained by using a 3D scanner, whereas the single-view point cloud was obtained by using multiple depth cameras. To unify the reference coordinate systems for the single-view point cloud and complete point cloud, an iterative closest point (ICP) algorithm was implemented to register the single-view point cloud to the complete point cloud coordinate system. Six independent objects and 50 single-view point clouds from the YCB and SSG datasets were respectively selected to enable object data accuracy comparison; the average reconstruction accuracies were calculated by applying different search radii.

The object data accuracy results are shown in Figure 10. The horizontal axis in the figure represents search radius ($r$) values; the red dashed line represents the average accuracy results for all objects, and the remaining curves show the average accuracy results for individual objects. The light green area between the red dashed line and the horizontal axis reflects the overall reconstruction accuracy. It can be seen that the overall accuracy of the object data generated from our SSG dataset was 6.47% higher than that generated as a result of using the YCB dataset; this proves that our method can be used to obtain higher quality object data. This is because the single-view point cloud for the YCB dataset was collected by using multiple depth cameras, and the high accuracy of the internal parameter calibration for each camera, as well as the coordinate transformation matrix required to synchronize camera data is difficult to guarantee. However, our method avoids such a complicated sensor calibration problem.

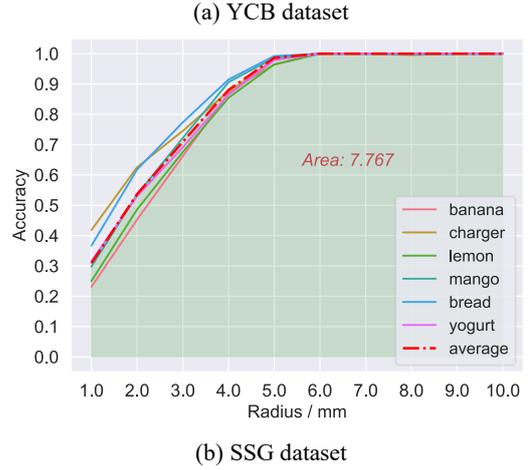

(a) YCB dataset

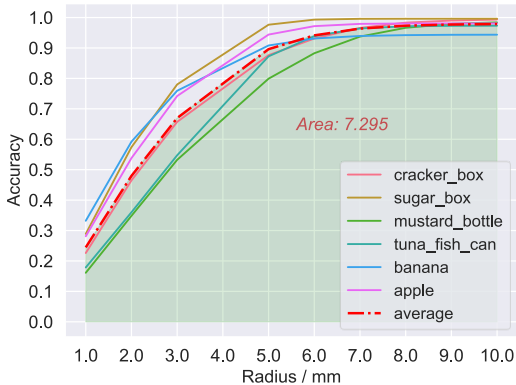

(b) SSG dataset

Figure 10. Object data accuracy results.

## 5.2 Simulated Experiment

To further verify the effectiveness of the proposed grasp data acquisition method, 1,500 grasp poses and their corresponding labels were generated for the aforementioned nine objects; nine objects from the YCB dataset were also selected for comparative analysis, and the same number of grasping poses and labels were generated for them. The experiment was carried out as follows: 1) Create a coefficient of friction list, i.e., $list_\mu$ = {3.0, 2.0, 1.7, 1.4, 1.3, 1.2, 1.1, 1.0, 0.9, 0.8, 0.7, 0.6, 0.5, 0.4, 0.3}; 2) Generate 100 grasp poses for each coefficient of friction in $list_\mu$; 3) To increase the discriminability between positive and negative samples, set thresholds $th_{good} = 0.45$ and $th_{bad} = 0.75$. Grasp poses with a $\mu$ less than or equal to $th_{good} = 0.45$ should be regarded as high-quality grasps, and grasp poses with a $\mu$ greater than or equal to $th_{good} = 0.45$ should be regarded as low-quality grasps. The labels were assigned according to the following formula:

$$label_g = \begin{cases} 1 & \mu_g \leq th_{good} \\ 0 & \mu_g \geq th_{bad} \end{cases} \quad (5)$$

To ensure equal numbers of positive and negative samples, 200 high-quality grasp poses for the nine objects and 200 randomly selected low-quality grasp poses were added to the dataset, with 20% being applied as the test set. Next, the geometric model of the gripper was used as a reference to process 50 single-view point clouds for the objects corresponding to each grasp pose, and the point cloud for the gripper closing area was extracted; the data were up-/down-sampled to 1,024 points. Thus, 180,000 sets of closing area point clouds and corresponding labels were generated for the two datasets, and subsequently used to train the grasp quality classification networks.

Three network models, GPD [3], PointNet [14], and PointNet++ [15], were implemented in the simulation experiment to classify the point cloud in the gripper closing area. Among them, PointNet and

PointNet++ are deep learning-based point cloud models, whereas GPD consists of conventional CNNs. The results of training each model on the YCB and SSG datasets are shown in Figure 11.

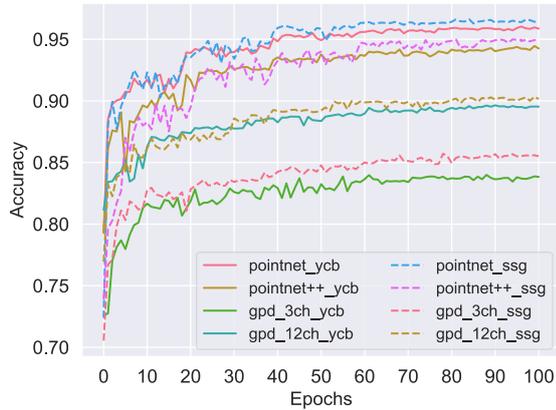

Figure 11. Accuracy of each tested classification model.

Analysis of the experimental results (Figure 11) revealed that the models trained on the SSG dataset was able to effectively learn, thereby proving the effectiveness of the proposed self-supervised learning-based method for manipulators. Additionally, the accuracy of each model trained on the SSG dataset tended to be higher than that of its counterpart trained on the YCB dataset. However, it should be noted that the test data were different; thus, it is not enough to prove that the proposed method can increase the accuracy of the classification model, and experiments must be carried out in real grasping scenes. It is also noteworthy that the classification accuracy achieved by applying a conventional GPD method to the two datasets was considerably lower than that obtained via the deep learning-based point cloud method; this confirms the feasibility of the proposed grasp quality classification method. Interestingly, the application of the proposed point cloud method to PointNet yielded the best classification results for the gripper closing area, even better than the improved version, PointNet++. This is because PointNet++ improved the ability of the network to extract localized point cloud information; this led to the network model paying too much attention to the localized point cloud information. This was a problem because the spatial domain of the point cloud for the gripper closing area was small; this means that there was not much localized information. Thus, the overall characteristics of the point cloud are more useful for classification.

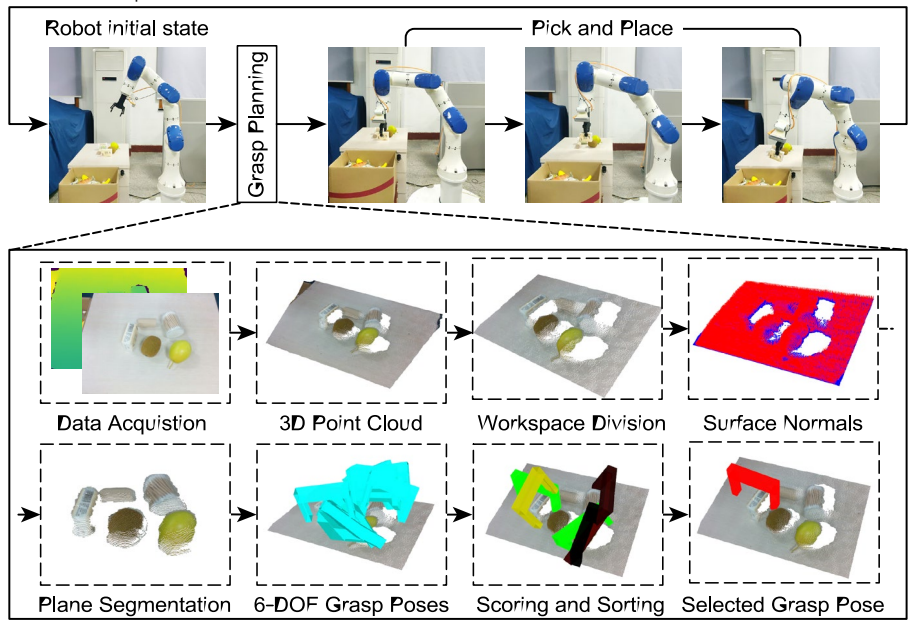

Figure 12. Grasping pipeline.

### 5.3 Grasp Experiment

Our grasping pipeline is shown in Figure 12. To ensure that that grasp task is performed with high efficiency and a high success rate, we applied several preprocessing procedures to the data from the depth sensor. First, the depth range was limited to 0–0.6. Then, instead of applying the point cloud format in ROS, the depth images and internal parameters of the camera were used to generate organized point clouds to increase the speed of calculation of the surface normal of the point cloud. Next, the workspace was setup to only take points within the following ranges as input: -0.2 m < x < 0.2 m and -0.2 m < y < 0.5 m. Finally, the RANSACE algorithm was applied to segment the desktop; the grasp points were uniformly sampled in the point cloud above the desktop.

(1) Single-target Grasp Experiment

In the single-target grasp experiment, only single-object data from the YCB and SSG datasets were used to train the grasp quality classification models, and five sets of grasp experiments were carried out for each trained model. Each set of experiments involved 36 grasps and nine objects from three different categories. A grasp was deemed to be successful if the 6-DOF grasp pose data output by the grasping planning algorithm resulted in the manipulator being able to grasp, lift, and hold the target object for 2 s while not touching other objects. The experimental results are summarized in Table 1; detailed results for one set of experiments are provided in Table 2. It can be seen that the models trained on our SSG dataset tended to outperform those trained on the YCB dataset; these results prove that the proposed method can increase the rate of successful grasps for a manipulator employed in a real scene. It should also be noted that, in the grasp experiments using fruit and other geometrically complex objects, the proposed PointNet-based method has a greater improvement than the conventional 6-DOF grasp planning method.

Table 1. Success rates of single-target grasp experiments.

| Dataset | Model | Fruit | Box | Column | Avg |
|---|---|---|---|---|---|
| YCB | GPD (12ch) | 63.3% | 68.3% | 73.3% | 68.3% |
| | PointNet | 81.6% | 86.7% | 91.7% | 86.6% |
| | PointNet++ | 75.0% | 81.7% | 90.0% | 82.2% |
| SSG | GPD (12ch) | 61.7% | 70.0% | 78.3% | 70.0% |
| | PointNet | 85.0% | 90.0% | 93.3% | **89.4%** |
| | PointNet++ | 78.3% | 83.3% | 91.6% | 84.4% |

(2) Multi-target Grasp Experiment

To further verify the advantages of the proposed self-supervised learning-based method for the manipulator, SSG data for various combinations of objects were added to the training set, and the above-mentioned three network models were retrained for grasp experiments in multi-target and cluttered scenes. Six objects were selected as the grasp targets, and 20 sets of experiments were performed using the models trained on the YCB dataset and our SSG dataset. The goal of each set of experiments was to perform a grasp task 10 times as the time is recorded. The experimental results are summarized in Table 3. The rate of

Table 2. Detailed experimental data for a set of single-target grasp tasks.

| Categories | Objects | PointNet | | | | | PointNet++ | | | | | GPD (12 ch) | | | | |
|---|---|---|---|---|---|---|---|---|---|---|---|---|---|---|---|---|
| | | Trials | | | | SR | Trials | | | | SR | Trials | | | | SR |
| | | #1 | #2 | #3 | #4 | | #1 | #2 | #3 | #4 | | #1 | #2 | #3 | #4 | |
| Fruit | Carambola | ✓ | ✓ | ✗ | ✓ | 3/4 | ✗ | ✓ | ✓ | ✗ | 2/4 | ✗ | ✓ | ✗ | ✓ | 2/4 |
| | Kiwi | ✓ | ✓ | ✓ | ✓ | 3/4 | ✓ | ✗ | ✓ | ✓ | 3/4 | ✓ | ✓ | ✓ | ✓ | 4/4 |
| | Pear | ✗ | ✓ | ✓ | ✓ | 3/4 | ✓ | ✓ | ✓ | ✓ | 4/4 | ✗ | ✓ | ✗ | ✗ | 1/4 |
| | Overall | | | | | 83.3% | | | | | 75.0% | | | | | 58.3% |
| Box | Pill box | ✓ | ✓ | ✓ | ✓ | 4/4 | ✓ | ✓ | ✓ | ✗ | 3/4 | ✗ | ✓ | ✗ | ✓ | 2/4 |
| | Paper napkin | ✓ | ✓ | ✓ | ✓ | 4/4 | ✓ | ✓ | ✓ | ✓ | 4/4 | ✗ | ✓ | ✓ | ✓ | 3/4 |
| | Wafer | ✓ | ✗ | ✓ | ✓ | 3/4 | ✓ | ✓ | ✗ | ✓ | 3/4 | ✓ | ✓ | ✗ | ✓ | 3/4 |
| | Overall | | | | | 91.6% | | | | | 83.3% | | | | | 66.7% |
| Column | Cylinder | ✓ | ✓ | ✓ | ✓ | 4/4 | ✓ | ✓ | ✓ | ✓ | 4/4 | ✓ | ✓ | ✓ | ✗ | 3/4 |
| | USB cable | ✗ | ✓ | ✓ | ✓ | 3/4 | ✗ | ✓ | ✓ | ✓ | 3/4 | ✓ | ✓ | ✗ | ✗ | 2/4 |
| | Cylindrical block | ✓ | ✓ | ✓ | ✓ | 4/4 | ✓ | ✓ | ✓ | ✓ | 4/4 | ✓ | ✓ | ✓ | ✓ | 4/4 |
| | Overall | | | | | 91.6% | | | | | 91.6% | | | | | 75.0% |
| Overall | | | | | | 88.9% | | | | | 83.3% | | | | | 66.7% |

Each model was trained on the SSG dataset; SR indicates the rate of successful grasps.
✓: the trial was successful  ✗: the trial was not successful  ✗: none of the grasps were kinematically feasible successful (SR) grasps in the table refers to the ratio of the number of successful grasps to the total number of grasps after the desktop has been emptied or 10 attempts have been made. The completion rate (CR) refers to the ratio of the number of removed objects to the total number of objects after 10 grasp attempts, and the grasp preparation time (PT) refers to the time spanning the beginning of image acquisition to the determination of the optimal grasp.

Table 3. Multi-target and cluttered scene grasp experiment results.

| Model | SR | | CR | | PT |
|---|---|---|---|---|---|
| | YCB | SSG | YCB | SSG | |
| GPD (12ch) | 64.8% | 66.6% | 79.2% | 82.5% | 1.22 s |
| PointNet | 82.6% | **86.2%** | 87.5% | **92.5%** | 1.37 s |
| PointNet++ | 79.3% | 82.9% | 85.0% | 90.8% | 3.43 s |

The experimental results show that, in terms of the rates of success and completion, the proposed 6-DOF grasp planning algorithm is best utilized by the PointNet or PointNet++ network model for multi-target and cluttered scene grasp tasks; furthermore, comparison to the conventional 6-DOF grasp planning method confirmed significant improvements. However, it is important to note that the implementation of the proposed algorithm in the PointNet++ network model significantly reduces grasp planning efficiency, making it unsuitable for grasp tasks with stringent real-time constraints. In addition, the models trained on our SSG dataset outperformed those trained on the conventional YCB dataset; this proves that the proposed self-supervised learning-based method can achieve a higher rate of successful grasps for manipulators applied in multi-target and cluttered scenes.

6. CONCLUSION

Unknown objects in an unstructured environment have various shapes and sizes. To ensure the robustness of the 6-DOF grasp planning algorithm for the manipulator, we developed a self-supervised learning mechanism. In this study, it was demonstrated to autonomously guide the manipulator to collect and label data when applied for use in real grasping scenes, eliminating the need for actual grasping operations. We also defined an object data accuracy index to quantitatively evaluate the quality of object datasets. The experimental results revealed that the proposed method can be employed to obtain high-quality grasp data, increase the rate of successful grasps for manipulators in real scenes, reduce the computational cost of grasp data acquisition, and realize a self-learning-based grasp planning method for manipulators. In the future, we plan to optimize the grasp planning module to further increase the grasp efficiency and success rate.

**Author Biographies**

**Peng Gang** (1973-) received the doctoral degree from the Department of control science and engineering of Huazhong University of Science and Technology (HUST) in 2002. Currently, he is an associate professor in the Department of Automatic Control, School of Artificial Intelligence and Automation, HUST. He is also a senior member of the China Embedded System Industry Alliance and the China Software Industry Embedded System Association, a senior member of the Chinese Electronics Association, and a member of the Intelligent Robot Professional Committee of Chinese Association for Artificial Intelligence. His research interests include intelligent robots, machine vision, multi-sensor fusion, machine learning and artificial intelligence.

**Zhenyu Ren** (1996-) received his bachelor degree in school of automation from Hainan University, China, in 2018. He received master's degree from the Department of Automatic Control, School of Artificial Intelligence and Automation, Huazhong University of Science and Technology. His research interests include intelligent robots and perception algorithms.

**Hao Wang** (1996-) received his bachelor degree in school of automation from Wuhan University of Technology, Wuhan, China, in 2019. He is currently a graduate student at the Department of Automatic Control, School of Artificial Intelligence and Automation, Huazhong University of Science and Technology, Wuhan, China. His research interests are intelligent robots and perception algorithms.

**Xinde Li** (1975-) received his doctoral degree in control theory and control engineering from the Department of Control Science and Engineering, Huazhong University of Science and Technology (HUST) in 2007. Thereafter, he joined the School of Automation, Southeast University, Nanjing, China, where he is currently a Professor and a Ph.D. Supervisor. From January 2012 to January 2013, he was a national public visiting scholar at Georgia Tech Visit and Exchange for one year. He was selected as an IEEE Senior member in 2016. From January 2016 to the end of August 2016, he worked as a Research Fellow in the ECE Department of the National University of Singapore. His main research interests include intelligent robots, machine vision perception, machine learning, human–computer interaction, intelligent information fusion, and artificial intelligence.